\documentclass[10pt,twocolumn,letterpaper]{article}

\usepackage[pagenumbers]{style}

\usepackage{cuted}
\usepackage{multirow}
\usepackage{makecell}
\usepackage{bm}

\definecolor{linkblue}{rgb}{0.21,0.49,0.74}
\usepackage[pagebackref,breaklinks,colorlinks,allcolors=linkblue]{hyperref}

\title{Beyond Prompt Imitation: Capability-Aware\texorpdfstring{\\}{ }Video Generation with Smaller Models}

\author{Ziqi Ding\textsuperscript{1,*} \quad Yuxin Cao\textsuperscript{2,*} \quad Kai Ye\textsuperscript{3} \quad Yaobo Yang\textsuperscript{1} \quad Wei Song\textsuperscript{4,\dag}\\[0.5em]
\textsuperscript{1}University of New South Wales, Australia\\
\textsuperscript{2}National University of Singapore, Singapore\\
\textsuperscript{3}The University of Hong Kong, Hong Kong SAR, China\\
\textsuperscript{4}Griffith University, Australia\\[0.3em]
{\small \textsuperscript{*}Equal contribution \quad \textsuperscript{\dag}Corresponding author}}

\begin{document}
\maketitle

\begin{abstract}
Inference-time methods that lift a small open video generator toward a proprietary one are compared through preference scores, which never say how much of the gap to a target video they close. We introduce the oracle control ladder, which hands a fixed student progressively richer information from a target clip and reads the closure of each fidelity metric at each rung as an empirical headroom reference for the methods that must synthesize that information. On 954 clips of the principal student, a dense caption reproduces little of the appearance and almost none of the motion of a target, a first frame closes half of the appearance gap and a tenth of the motion gap, a depth video closes three quarters of the motion gap and leaves color untouched, and the two together close 64 to 94\% of every fidelity gap across four teachers, while a second student agrees on the rungs that it accepts. An item response model forecasts which rung closes which clip with held-out AUC of 0.76 to 0.95 on the fidelity criteria and serves as the prior of a budgeted selection policy. When controls are free, a fixed dense-control rung is hard to improve upon, and once controls are priced the policy lies above the fixed-rung frontier at matched cost and survives critics that ruin critic-driven search.
\end{abstract}

\section{Introduction}
\label{sec:intro}

\begin{figure*}[t]
\centering
\includegraphics[width=\textwidth]{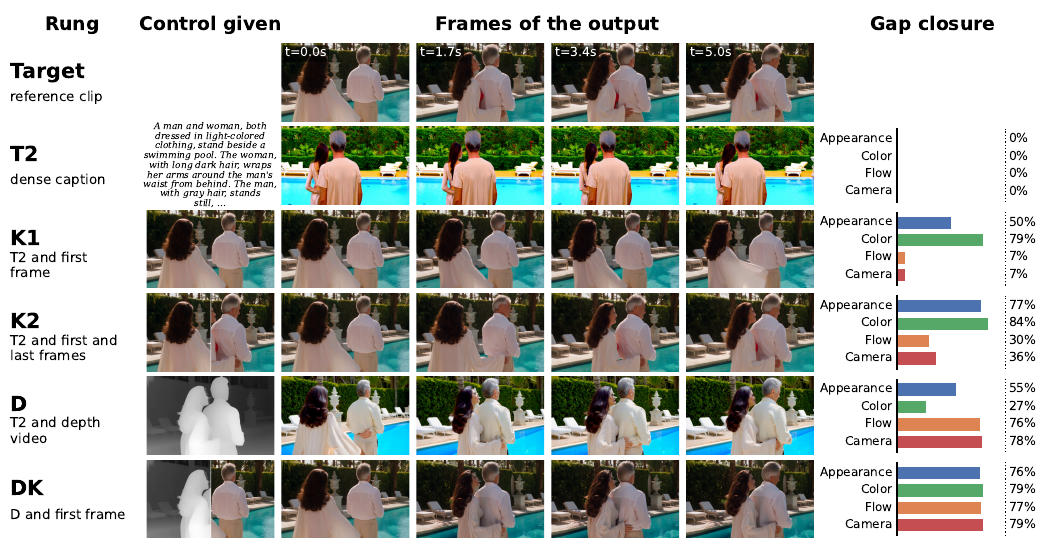}
\caption{The oracle control ladder on one target clip. Each rung hands the student the control in the second column, and the bars give the share of the gap to the target that the rung closes on four criteria. The caption reproduces the scene and little else, a first frame fixes the appearance and leaves the motion open, a depth video fixes the motion and leaves the appearance open, and the two together fix both.}
\label{fig:teaser}
\end{figure*}

Video generation has split into two tiers. Proprietary systems such as Sora, Veo 3 and Kling produce clips whose motion, framing and event structure most users cannot obtain from an open model that runs on one GPU, while open models of one to five billion parameters~\cite{wan,cogvideox,ltxvideo} are the ones that users can run, adapt and control~\cite{ppnas}. Many methods promise to lift a small model toward a large one at inference time~\cite{splitinference,medshield}. Prompt optimizers rewrite the  text~\cite{rapo,vpo,promptavideo}, agents critique and regenerate~\cite{vista,phyt2v,videorepair}, search methods spend more samples~\cite{itscaling,videot1}, and control methods hand the model keyframes, depth or trajectories~\cite{vace,sparsectrl,motionprompting}. Each is evaluated against a text prompt and a preference score, so the literature says which method scores higher but never how much of the gap to a specific target a given student can close with a given kind of information, or which part of that gap it can close at all.

This paper asks the question directly. Given a reference clip that a user wants to reproduce and a fixed small student, we measure how far the student gets when it is told progressively more about the reference. A planner reads the reference and passes the student a short caption, a dense caption, a structured program, true keyframes, the true depth video or the true pose video. We call this sequence the oracle control ladder, because the planner sees the target, which no prompt optimizer or structure planner can do, so the closure at each rung is an empirical headroom reference, conditioned on privileged information about the target, for the methods that use a comparable interface with this student. The reference reflects an advantage of information access under a fixed student and sampler, and a method that changes the architecture, the training or the sampler is measured against it only through the information that it hands over. Figure~\ref{fig:teaser} shows the ladder on one clip. If the dense caption closes a tenth of the appearance gap and the first frame closes most of it, then prompt optimization that keeps this student has about a tenth of the gap to work with, and the useful work lies in producing one good frame.

Three properties make the measurement informative. Firstly, the gap is measured per metric against the reference itself, with the reference's own score as the ceiling, so the closure of appearance, structure, motion, camera, events and quality can be read separately. Secondly, the closure varies across clips in a way that clip features explain~\cite{fedward}. We fit a many-facet Rasch model~\cite{rasch1960,fischer1973} in which each rung has an ability per criterion and each clip has a difficulty that its motion, subjects and length predict, so the model forecasts for an unseen clip which information a student can turn into fidelity, and it profiles a new student with a handful of adaptive probes. Thirdly, the fitted model is a prior for the decision that a user faces under a budget, namely which rungs to try and when to stop. We cast that decision as a Pandora's box problem~\cite{weitzman1979}, whose reservation-value rule is optimal for costly inspection, and we replay it on the full grid against Best-of-K sampling, uniform compute and fixed strategies under critics of decreasing reliability, since adaptive policies collapse when the verifier is weak~\cite{computeaudit}.

We run the ladder on 954 reference clips from four proprietary generators and three consumer tools with Wan2.1-VACE-1.3B as the main student, and we report every finding with bootstrap intervals over clips. A second student with four times the parameters runs the rungs that its interface accepts, the teacher clips carry the content dimension of their prompt, and a two-round prompt refinement loop with the same planner as its critic places one practical method of the first family against the reference that the ladder gives it. We also validate the composite critic against the judgment of 10 human raters on pairs of outputs and check that the sampling schedule of the grid preserves every closure that we report. The appendix details the clips and settings and shows every rung's outputs on four clips chosen by a fixed rule.

Our contributions are as follows.
\begin{itemize}
\item We introduce the oracle control ladder, which feeds a fixed student progressively more of a target clip and reads the gap closure at each rung as an empirical headroom reference for methods with a comparable interface.
\item We fit an explanatory item response model with rungs as facets and feature-predicted clip difficulties, which forecasts closure on unseen clips with readable parameters and profiles a new student from few probes.
\item We cast budgeted selection as a Pandora's box problem with the capability model as its prior, replay it at matched control cost and under weak critics, and release the generated grid, the scores and the models.
\end{itemize}

\section{Related Work}
\label{sec:related}

\noindent\textbf{Prompt Optimization and Agentic Refinement.} The first family changes only the text input. Promptist~\cite{promptist} learned prompt adaptation for images, and RAPO~\cite{rapo}, VPO~\cite{vpo} and Prompt-A-Video~\cite{promptavideo} carried the idea to video. A second generation closes the loop with the output~\cite{phyt2v,rapopp,vqqa,vista,pris}, and agentic systems plan layouts or run design, generation and verification loops~\cite{videodirectorgpt,lvd,genmac,coagent,vgot,mora,videorepair}. None of them measures what a prompt achieves for a specific target video with a fixed student, which is the reference that our oracle control ladder gives.

\noindent\textbf{Controllable and Reference-Driven Generation.} Control signals carry information that text cannot. Video diffusion models accept depth, edges, sketches and sparse keyframes~\cite{videocomposer,controlvideo,sparsectrl,ctrladapter}, trajectories and camera paths~\cite{motionctrl,cameractrl,tora,motionprompting,gowiththeflow,das} and keyframes to interpolate~\cite{geninbetween,fcvg}, and unified models such as VACE~\cite{vace}, FullDiT~\cite{fulldit} and Wan~\cite{wan} take most conditions in one network~\cite{ctrlsurvey}. Reference-driven methods take the information from an existing video by fine-tuning on one clip~\cite{motiondirector,vmc}, steering attention or flow at sampling time~\cite{stdf,motionclone,ditflow,conmo,flowmotion}, preserving a subject~\cite{phantom,saber,videoasprompt} or following a stronger model~\cite{videoguide}. Each method fixes one way of using the reference, and we treat them as rungs of one ladder.

\noindent\textbf{Evaluation of Generated Video.} Benchmarks decompose quality into dimensions that metrics score and human preference validates~\cite{vbench,evalcrafter,t2vcompbench,videobench,qeval}, VBench-2.0~\cite{vbench2} adds videos from commercial generators that we reuse as teachers, question-based judges ask a VLM about the output~\cite{etva,evalagent,videoscore}, and physics benchmarks test commonsense~\cite{videophy,physicsiq}. FVD~\cite{fvd} favors per-frame quality~\cite{fvdbias}, and dynamics are measured separately~\cite{devil}. Reference-based evaluation is rarer~\cite{ref4d,wyd}, and our metrics pair every score with the self-score of the reference to define gap closure.

\noindent\textbf{Inference-Time Scaling and Stopping.} Sampling-time compute improves diffusion models through noise search, verifier-guided selection and particle methods~\cite{itscaling,evosearch,videot1,ppautodl}, repeated sampling obeys scaling laws~\cite{monkeys,snell}, and adaptive allocation learns how much to spend per input~\cite{howhard}, but sampling headroom and selection gain diverge for video~\cite{computeaudit}. Stopping rules trade accuracy against cost~\cite{verifierthreshold,stoppingvsbon}, and the Pandora's box problem~\cite{weitzman1979} gives the optimal order and stopping time for costly inspection through reservation values~\cite{gittins,pandoragittins,pandorarouting}. We apply it to the choice of what information to give a generator.

\begin{figure*}[t]
\centering
\includegraphics[width=\textwidth]{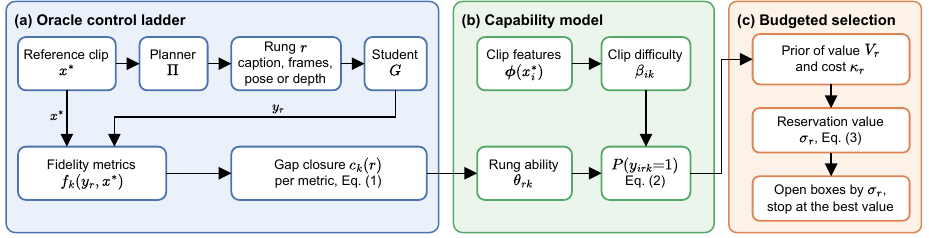}
\caption{Overview. The ladder (a) reads the gap closure of each rung of reference information, the capability model (b) predicts it from clip features, and budgeted selection (c) opens rungs by their reservation values.}
\label{fig:overview}
\end{figure*}

\noindent\textbf{Item Response Theory.} IRT models the probability of a correct answer through ability and difficulty~\cite{rasch1960}, and explanatory models express difficulty through item features~\cite{fischer1973}. Language processing uses IRT to build scales~\cite{lalor2016}, compress benchmarks~\cite{tinybench,metabench}, run adaptive tests~\cite{fluid,atlas}, route queries~\cite{irtrouter} and beyond~\cite{trustirt,lostbench,mmirt,generalscales}. In visual generation, FlashEval~\cite{flasheval} selects evaluation sets, prompt performance prediction~\cite{promptperf} forecasts image quality from the prompt, and TasteRoute~\cite{tasteroute} routes video prompts among generators. We bring explanatory IRT to video generation with the information given to a fixed student as the facet, so the fitted parameters say which clips a rung helps.

\section{Oracle Control Ladder}
\label{sec:ladder}

\noindent\textbf{Setting.} Figure~\ref{fig:overview} summarizes the three components of the paper. A reference clip $x^\ast$ shows what a user wants, and a small student generator $G$ takes a text prompt and, depending on its interface, frames, a dense control video and reference images. A planner $\Pi$ reads the reference and extracts the information that the student receives. We call each choice of extracted information a rung $r$ and write $y_r = G(\Pi_r(x^\ast), s)$ for the student output under rung $r$ and seed $s$, and the ladder orders the rungs by how much of the reference they expose. Because the planner sees the target itself, every rung is an oracle for the method family that would have to recover the same information from the user's prompt and from its own outputs. A prompt optimizer cannot exceed the best text rung in information, a keyframe pipeline cannot exceed the rung that receives the true anchor frames, and a structure planner cannot exceed the rung that receives the true depth, so for a fixed student and inference configuration the closure at a rung is an empirical headroom reference for the methods with that interface, conditioned on privileged information about the target.

\noindent\textbf{Rungs.} Table~\ref{tab:rungs} lists the ten rungs. The short caption is one sentence, the dense caption describes subjects, scene, camera, timeline and style in 80 to 120 words, and the program renders the same content as a structured list of fields. The keyframe rungs pass one, two or three true frames, and the student inpaints the rest. The depth rung passes a per-frame relative depth video, the two dense rungs combine depth with the first frame or with a reference image of the subject, and the pose rung passes whole-body skeletons on the clips with a visible person. Every rung above the first uses the dense caption as its prompt, so the rungs differ only in the visual information that they add, and Table~\ref{tab:rungs} orders them by the compressed size of the control.

\begin{table}[t]
\centering
\caption{Rungs of the ladder and their median input size (kB).}
\label{tab:rungs}
\small
\setlength{\tabcolsep}{4pt}
\begin{tabular}{lp{2.55cm}p{2.75cm}r}
\toprule
Rung & Information given to the student & Family whose headroom it references & Size \\
\midrule
T1 & short caption & prompt rewriting & 0.1 \\
T2 & dense caption & prompt optimization & 0.3 \\
T3 & structured program & planning agents & 0.3 \\
K1 & T2 and first frame & image-to-video pipelines & 70 \\
P & T2 and pose video & character animation & 105 \\
K2 & T2 and first and last frames & keyframe pipelines & 140 \\
K3 & T2 and three keyframes & storyboard pipelines & 210 \\
D & T2 and depth video & structure planners & 851 \\
DK & D and first frame & dense video-to-video transfer & 920 \\
DR & D and a subject reference image & subject-driven structure planners & 920 \\
\bottomrule
\end{tabular}
\end{table}

\noindent\textbf{Metrics and Gap Closure.} Each output is compared with its reference by fidelity metrics $f_k(y, x^\ast) \in [0, 1]$ that reach 1 when the output equals the reference, and it is also scored by no-reference quality metrics $q_k(y)$ that the reference itself can be scored with, which Section~\ref{sec:setup} lists. The dense caption rung T2 is the base of the ladder, because it is the strongest condition that every text-to-video student accepts, and the gap closure of rung $r$ on metric $k$ is
\begin{equation}
c_k(r) = \frac{f_k(y_r, x^\ast) - f_k(y_{\mathrm{T2}}, x^\ast)}{1 - f_k(y_{\mathrm{T2}}, x^\ast)},
\label{eq:closure}
\end{equation}
and the no-reference form replaces the ceiling 1 by the score of the reference, $q_k(x^\ast)$. A closure of 1 means that the rung makes the student match the reference on that metric, 0 means that the information made no difference, and a negative closure means that it hurt. Closure is a paired quantity, computed per clip and seed before averaging, so every confidence interval that we report comes from a bootstrap over the clips of the same set. Closure measures reference fidelity, the reproduction of a specified target, and a higher closure does not imply a better video by a no-reference standard, which Section~\ref{sec:res-ladder} reports separately.

\section{Capability Model and Budgeted Selection}
\label{sec:method}

\noindent\textbf{Capability as an Item Response.} The ladder produces, for every clip $i$, rung $r$ and criterion $k$, a success indicator $y_{irk}$ that equals 1 when the closure $c_k(r)$ on clip $i$ reaches a threshold $\tau$. We model the success probability with a many-facet Rasch model in which the rung is a facet of the respondent and the clip is the item,
\begin{equation}
\log \frac{P(y_{irk} = 1)}{P(y_{irk} = 0)} = \theta_{rk} - \beta_{ik},
\label{eq:irt}
\end{equation}
where $\theta_{rk}$ is the ability of rung $r$ on criterion $k$ and $\beta_{ik}$ is the difficulty of clip $i$ on that criterion, so the vector $\theta_{r\cdot}$ is the capability profile of a rung. Difficulty is explained by clip features in the manner of the linear logistic test model~\cite{fischer1973}, $\beta_{ik} = b_k + \mathbf{w}_k^{\top}\boldsymbol{\phi}(x^\ast_i)$, with features $\boldsymbol{\phi}$ such as the magnitudes of camera and object motion, the number of subjects, the presence of a person, the aesthetic score and warping error of the reference, and the lengths of the caption and the clip, so the model predicts the difficulty of a clip that it has never seen. We fit the model by maximum likelihood with a small ridge penalty and compare it on held-out clips with a per-rung success rate and with a gradient-boosted classifier trained on the same clip features.

\noindent\textbf{Adaptive Profiling of a New Student.} Fitting Equation~\eqref{eq:irt} on the full grid of one student calibrates the clip difficulties, and a new student is then profiled with few generations by adaptive testing, where each probe selects the clip whose outcome carries the largest expected information about the abilities of the new student.

\noindent\textbf{Budgeted Selection as a Pandora's Box.} Given a new reference and a budget, the practical question is which rungs to try and when to stop. A box is a rung $r$ at one sampling seed, with cost $\kappa_r = t_r + \nu s_r$, where $t_r$ is the time of generation and control extraction, $s_r$ is the size of the control and $\nu$ prices the information itself. Its value $V_r$ is the composite closure of the output, namely the mean of the closures of Equation~\eqref{eq:closure} on the seven fidelity criteria and the checklist, each clipped to [-1, 1], which is uncertain until the box is opened. The composite mixes the reference-based closures with the checklist closure because the named events are part of the target that the user wants to reproduce, and the no-reference scores stay outside the value. The prior of $V_r$ comes from the clip features of the capability model through a regression per rung fitted on the other folds, every seed of a rung is an independent draw from it, opening a box reveals its value through the critic, and the user keeps the best output found. This is the Pandora's box problem of Weitzman~\cite{weitzman1979}, whose optimal policy assigns each box a reservation value $\sigma_r$, the sure value at which a searcher is indifferent between taking it and opening the box, and the reservation value solves the equation
\begin{equation}
\mathbb{E}\!\left[\max(V_r - \sigma_r, 0)\right] = \lambda \kappa_r,
\label{eq:reservation}
\end{equation}
where $\lambda$ is the price of one second in units of closure, opens boxes in decreasing order of $\sigma_r$, and stops as soon as the best value found exceeds the largest reservation value of the boxes that remain. The rule is optimal when the values are independent given the features and the critic is exact, and neither holds exactly, so we test the policy by replay on the full grid, where every box of every clip has been generated and the policy pays only for the boxes that it opens, against Best-of-K sampling, uniform opening, every fixed rung and an oracle, with a random order, a greedy order by the prior mean and a per-clip routing rule that opens the single rung with the largest predicted value net of its priced cost as ablations and baselines. The replay estimates the outcome of a policy on a grid that was generated once in advance, so every policy sees the same boxes and the same critic, and it is a cost-sensitive analysis of selection on that grid that does not demonstrate an end-to-end deployment.

\section{Experiments}
\label{sec:experiments}

\subsection{Experimental Setup}
\label{sec:setup}

\noindent\textbf{Reference Clips.} The references come from 61 user videos made with Seedance, Jimeng and Kling and from the sampled-video release of VBench-2.0~\cite{vbench2}, with the same 204 prompts for each of four teachers, Kling, Sora, Veo 3 and Vidu Q1. Every video is cut into shots, and each shot yields clips of 81 frames at 16 fps, or 49 frames when the shot is shorter than 5 s, at about 480p. The user videos give 148 clips and the teacher videos 806, for 954 reference clips in total, and Section~\ref{sec:app-clips} in the appendix gives the details.

\noindent\textbf{Student, Planner and Judge.} The student is Wan2.1-VACE-1.3B~\cite{wan,vace}, whose interface accepts a text prompt, kept frames, a dense control video and reference images, so every rung of Table~\ref{tab:rungs} runs on the same weights, with the UniPC scheduler for 30 steps at a guidance scale of 5.0, validated against 50 steps in Section~\ref{sec:app-steps} in the appendix. All experiments run on NVIDIA RTX PRO 5000 GPUs with 48 GB. A second student, Wan2.2-TI2V-5B~\cite{wan}, with four times the parameters and a different transformer and VAE, accepts a text prompt with an optional first frame and runs the caption rungs and K1 on the same clips and seeds. The planner Qwen3-VL-8B-Instruct reads 12 timestamped frames and writes the three text rungs and the checklist questions, depth comes from Depth Anything V2~\cite{depthanything2} and pose from DWPose~\cite{dwpose}, and the judge that answers the checklist is InternVL3.5-8B~\cite{internvl35}, from a different family than the planner that writes the questions.

\noindent\textbf{Baselines and Replay.} We run one specific prompt refinement protocol, a loop in the manner of PhyT2V~\cite{phyt2v}, RAPO++~\cite{rapopp} and VISTA~\cite{vista} with the planner as the critic, for two rounds on the 148 user clips, and we report the best of its candidates and the best of as many seeds of the dense caption under the composite critic of Section~\ref{sec:method}. The budgeted replay runs on the same clips with five seeds of the dense caption, three of K2 and DK and one of K1 and D, with measured times and control sizes as costs, a ridge prior on the clip features and critics that are exact, noisy, appearance-only or no-reference. Section~\ref{sec:app-impl} in the appendix gives every setting of the replay.

\noindent\textbf{Metrics.} Fidelity metrics compare an output with its reference on frames sampled at a stride of 5. Appearance (DINO) is the mean cosine similarity of DINOv2~\cite{dinov2} class tokens of corresponding frames, structure (Struct) the same with their patch tokens, and semantics (CLIP) the same with CLIP ViT-L/14~\cite{clip} embeddings. Depth is the correlation of standardized depth maps, and Color is one minus half the $L_1$ distance of RGB histograms. Motion metrics use RAFT~\cite{raft} optical flow. Flow is the normalized agreement $1 - \mathrm{EPE} / (\lVert u \rVert + \lVert v \rVert)$ of the two flow fields, where EPE is the end-point error, Camera the same agreement of a per-frame translation and zoom fitted to each field, and Object the agreement of the residual flows. The checklist score (Check) averages the fraction of the planner's yes-questions that the judge confirms on the output and the fraction of its no-questions that the judge rejects, over the questions that the judge answers correctly on the reference itself. No-reference metrics are the LAION aesthetic score~\cite{laion5b} (Aes), subject and background consistency as in VBench~\cite{vbench}, the warping error and the dynamic degree. Scoring the reference against itself gives the ceiling of Equation~\eqref{eq:closure}.

\subsection{The Oracle Control Ladder}
\label{sec:res-ladder}

\begin{table}[t]
\centering
\caption{Scores at the dense caption rung and gap closure (\%) of the other rungs on the user clips.}
\label{tab:ladder}
\small
\setlength{\tabcolsep}{3pt}
\begin{tabular}{lrrrrrrrrrr}
\toprule
 & T2 score & T1 & T3 & K1 & P & K2 & K3 & D & DK & DR \\
\midrule
DINO & 0.53 & -39 & -11 & 53 & 19 & 73 & 83 & 50 & 78 & 61 \\
CLIP & 0.68 & -20 & -1 & 50 & 5 & 67 & 76 & 21 & 63 & 33 \\
Struct & 0.31 & -5 & -1 & 30 & 23 & 48 & 64 & 47 & 68 & 53 \\
Depth & 0.48 & -14 & -13 & 36 & 50 & 60 & 79 & 91 & 95 & 93 \\
Color & 0.38 & -9 & -4 & 55 & 6 & 68 & 76 & 7 & 61 & 13 \\
Flow & 0.18 & -9 & -5 & 8 & 49 & 29 & 45 & 73 & 76 & 74 \\
Camera & 0.21 & -17 & -12 & 5 & 52 & 28 & 46 & 75 & 77 & 74 \\
Object & 0.16 & -5 & -3 & 5 & 33 & 18 & 32 & 55 & 62 & 57 \\
Check & 0.88 & -99 & -3 & 61 & 35 & 76 & 83 & 51 & 69 & 66 \\
\bottomrule
\end{tabular}
\end{table}

Table~\ref{tab:ladder} reports the ladder on the 148 user clips at one seed, where pose applies to the 105 clips with a visible person, Figure~\ref{fig:ladder} in the appendix draws it with bootstrap intervals, and Figures~\ref{fig:examples-user} and \ref{fig:examples-teacher} in the appendix show the outputs on four clips. The dense caption leaves a large gap on every fidelity axis. Its outputs reach a DINO similarity of 0.53, a structure similarity of 0.31 and a flow agreement of 0.18 with the reference, so a prompted student reproduces little of the appearance and almost none of the motion of a given video, while the caption already satisfies 88\% of the validated event questions, which leaves little headroom on the semantic axis for prompt rewriting.

The two other text rungs sit below the dense caption. The short caption gives up 39\% of the DINO gap and 99\% of the checklist headroom, so the events that a prompt names are the first thing that a short prompt loses, and the structured program is below the dense caption by 11 to 13\% of the DINO, depth and camera gaps on either student, so a program written from the reference gains nothing over a well-written caption for this student.

The visual rungs split the gap along the axes, and the split is sharp. The first frame alone closes 53\% of the appearance gap and 55\% of the color gap but only 5 to 8\% of the motion gaps, so a keyframe tells the student what the video looks like and nothing about how it moves. Adding the last frame raises the flow and camera closure to 28 to 29\%, because the pair bounds the camera path, and three keyframes close 83\% of the appearance gap and 45 to 46\% of the motion gaps, so a storyboard pins the look of a video and half of its motion. Pose is the mirror image of a keyframe, with 49 to 52\% of the flow and camera gaps and 19\% of the appearance gap. Depth alone closes 91\% of the depth gap and 73 to 75\% of the flow and camera gaps but 7\% of the color gap, so a depth video tells the student how the scene moves and leaves it to guess what the scene looks like. A subject reference image recovers part of what depth lacks, since DR lifts the DINO closure from 50 to 61\%, but the first frame recovers far more, since DK closes 61 to 95\% of the gap on every fidelity axis. Object motion is the hardest axis, with 62\% closure even at DK, and its residual gap belongs to the capacity of the student.

Target fidelity and perceptual quality pull in different directions, and the trade-off is substantive. Every visual rung that fixes frames lowers the no-reference quality of the outputs. The aesthetic score falls by 0.07 at K1, 0.20 at K2, 0.24 at K3 and 0.15 at DK on its 10-point scale, and subject consistency falls by 0.05 to 0.07 on every visual rung, so the information that makes an output match its target constrains the student away from the outputs that it prefers, and a method judged by preference alone would not see the fidelity that it gains. The checklist closure of the visual rungs, 35\% at P to 83\% at K3, is lower than their closure on appearance, so dense structure carries the student through the right motion while the named events are rendered less reliably than the appearance of the scene.

The averages hide that clips differ in what they need, and Section~\ref{sec:app-suff} in the appendix reports the lowest rung that closes half of the gap on each clip. A first frame is enough for appearance on 59\% of the user clips, pose or depth is the lowest sufficient rung for flow on two thirds of them, and no rung reaches half closure on object motion for 22\%. Adding a keyframe moves the motion away from the reference on a third to a half of the clips even though it improves the mean.

\subsection{Teacher Clips and Content Dimensions}
\label{sec:teacher}

Table~\ref{tab:teacher} in the appendix runs the ladder on all 954 clips by the source of the reference, and Table~\ref{tab:dimension} in the appendix breaks the 806 teacher clips down by the VBench-2.0 dimension of their prompt, 17 dimensions of 44 to 48 clips. On all clips, the first frame closes 55\% of the appearance gap and 9\% of the flow gap, the depth video closes 70\% of the flow gap and 5\% of the color gap, and the pair closes 64 to 94\% of every fidelity gap. The four teachers agree on nearly every cell, with the first frame at 49 to 59\% of the appearance gap, the depth video at 63 to 72\% of the flow gap and the pair at 90 to 95\% of the depth gap. The dimension of a prompt moves the closure far more than its teacher does, along the axis that the dimension names. The depth video closes 37\% of the flow gap on Thermotics, 49\% on Material and 53\% on Dynamic Attribute, where the motion of the reference is a change of the surface itself that a depth map does not carry, and 80 to 83\% on Human Anatomy, Human Interaction and Motion Order Understanding.

\subsection{Prompt Refinement Against the Ladder}
\label{sec:refine}

\begin{table}[t]
\centering
\caption{Held-out AUC of the success predictors and fitted rung abilities at half closure.}
\label{tab:irt}
\small
\setlength{\tabcolsep}{4pt}
\begin{tabular}{lrrrrrrr}
\toprule
 & \multicolumn{3}{c}{AUC} & \multicolumn{4}{c}{Rung ability $\theta_{rk}$} \\
\cmidrule(lr){2-4}\cmidrule(lr){5-8}
 & Rate & Rasch & GBM & K1 & K2 & D & DK \\
\midrule
DINO & 0.73 & 0.76 & 0.75 & 0.5 & 1.9 & 0.1 & 3.2 \\
CLIP & 0.83 & 0.90 & 0.90 & 0.0 & 1.9 & -3.4 & 1.8 \\
Struct & 0.74 & 0.89 & 0.92 & -2.2 & -0.8 & -0.8 & 2.9 \\
Depth & 0.85 & 0.89 & 0.89 & -0.6 & 1.0 & 3.8 & 4.0 \\
Flow & 0.91 & 0.95 & 0.95 & -3.0 & -1.4 & 2.6 & 3.0 \\
Camera & 0.87 & 0.92 & 0.93 & -2.6 & -1.0 & 2.3 & 2.4 \\
Object & 0.86 & 0.90 & 0.92 & -4.0 & -2.7 & 0.7 & 1.1 \\
Check & 0.57 & 0.47 & 0.52 & 1.0 & 1.7 & 0.6 & 1.5 \\
\bottomrule
\end{tabular}
\end{table}

Table~\ref{tab:refine} in the appendix places the tested refinement protocol on the ladder. The first round raises the composite closure by 2.1\% over the dense caption, with a bootstrap interval of [0.0, 4.1]\% that touches zero, and the second round gives back 0.5\%, so the loop converges after one round at a level that the interval cannot separate from its starting point. The rounds move the semantic axis and little else. The checklist closure rises to 13\% after one round and to 17\% after two, while appearance moves by 3 to 5\% and the motion axes fall by 4 to 9\%, so the motion words of a rewritten prompt move the camera without moving it toward the motion of the reference.

A practical loop keeps the best of its candidates, which closes 8.0\% of the composite gap and 37\% of the checklist gap, while three seeds of the dense caption selected by the same critic close 7.3\% and 35\%, and both figures are optimistic because the critic that selects the candidate is the metric that scores it. Two results follow, and they stay apart. Firstly, the oracle text rungs leave most of the fidelity gap open, as Section~\ref{sec:res-ladder} measured. Secondly, under the evaluated two-round protocol, prompt rewriting yields 0.7\% of composite closure beyond seed selection at a matched generation count, and the best round stays below the first-frame rung on every fidelity axis, at 9\% against 53\% of the appearance gap. The second result describes this protocol, and a specialized prompt optimizer or a longer search is placed against the same reference only by running it.

\subsection{Second Student}
\label{sec:second}

Table~\ref{tab:wan22} in the appendix runs the caption rungs and K1 on Wan2.2-TI2V-5B over the same 148 user clips and seeds. The second student starts from almost the same place as the first, with a DINO similarity of 0.56 against 0.53 and a flow agreement of 0.19 against 0.18 at the dense caption, so four times the parameters do not move a prompted student closer to a given video. The first frame then closes the same share of the gap on both students, within 2\% on appearance, CLIP, structure, depth and color, and 12 to 13\% of the flow and camera gaps on the larger student against 5 to 8\% on the smaller, so a keyframe tells either student what the video looks like and tells neither how it moves. The caption rungs agree as well, with the program below the dense caption on every criterion of both students. On the rungs that both students accept, the share of the gap that a rung closes is therefore mostly a property of the information in the rung, and the student decides the few points that remain. The evidence for the depth and pose rungs comes from Wan2.1-VACE-1.3B alone, and whether the dense-control ladder transfers to other architectures remains open.

The capability model also profiles the second student from few generations. With the clip difficulties of the first student held fixed, the K1 abilities of Wan2.2 lie within 0.4 logits of those of VACE on every criterion, and adaptive probing brings the mean ability error below 0.5 logits after 7 probes where a random order of clips needs 18, as Section~\ref{sec:app-profile} in the appendix shows.

\subsection{Capability Model}
\label{sec:capability}

Table~\ref{tab:irt} evaluates the capability model of Equation~\eqref{eq:irt} on the success indicators at $\tau = 0.5$, with 592 clip-rung observations per fidelity criterion over 148 clips and five-fold cross-validation over clips, so every prediction is made for a clip that the model has not seen. The per-rung success rate (Rate) is the baseline. The Rasch model with explanatory difficulty raises the held-out AUC over that baseline by 0.04 to 0.15 on CLIP, structure, depth, flow, camera and object motion, and the gradient-boosted classifier (GBM) on the same features is within 0.02 of it on six of the eight criteria and ahead by 0.03 on structure, so the clip features carry real information about which clip will be closed by which rung, and the predictive range is metric-specific, from 0.76 on appearance to 0.95 on flow. The advantage of the Rasch form over the boosted classifier is an interpretable capability profile, parameters shared across rungs and the few-shot profiling of Section~\ref{sec:second}, since its held-out AUC is comparable. Every feature is read from the reference clip, its flow, its planner program and its self-scores, so the prior serves a user who holds the target, and a text-only deployment is untested. Appearance has the smallest gain, 0.03, because the first frame closes the DINO gap on most clips, and the checklist stays at chance, since the caption already satisfies most questions. The fit on all 954 clips in Table~\ref{tab:irtall} in the appendix keeps the gains on every fidelity criterion.

The fitted abilities are the capability profile of the student. DK has an ability above 2 on structure, depth, flow and camera, K1 is below -2 on structure, flow and camera, and the order reverses on CLIP, where depth alone has an ability of -3.4. The largest feature weights belong to the two motion magnitudes. Camera motion lowers the difficulty of structure and depth, with weights of -1.3 and -0.5, and raises that of flow and camera, with 0.9 and 0.8, while object motion lowers the difficulty of flow and camera with -1.5. Table~\ref{tab:weights} in the appendix lists every weight.

\subsection{Budgeted Selection}
\label{sec:selection}

\begin{figure*}[t]
\centering
\includegraphics[width=\textwidth]{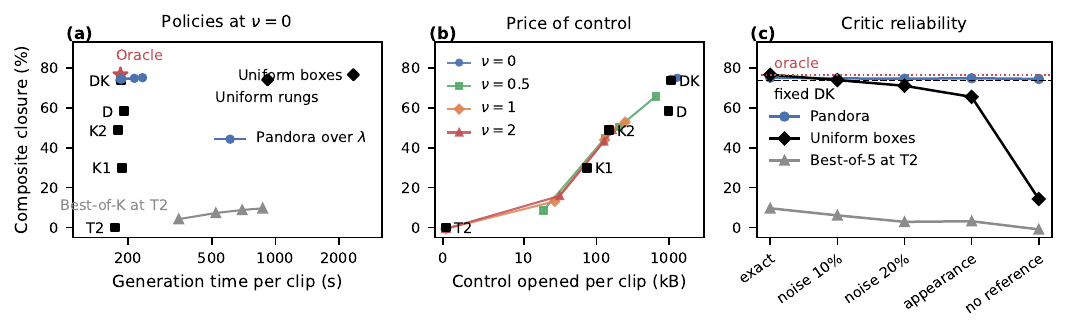}
\caption{Replay of the selection policies on the user clips. (a) Value against generation time at $\nu = 0$. (b) The Pandora policy over $\lambda$ at four prices of control. (c) Value under critics of decreasing reliability at $\nu = 0$.}
\label{fig:select}
\end{figure*}

\begin{table}[t]
\centering
\caption{Closure (\%) at matched control per clip once control is priced, against the fixed-rung frontier and the routing rule.}
\label{tab:matched}
\small
\setlength{\tabcolsep}{4pt}
\begin{tabular}{rrrrrr}
\toprule
$\nu$ & $\lambda$ & Control (kB) & Policy & Frontier & Routing \\
\midrule
0.5 & 0.0005 & 656 & 65.7 & 54.6 & 65.8 \\
0.5 & 0.001 & 206 & 50.6 & 49.6 & 50.4 \\
0.5 & 0.002 & 151 & 48.6 & 49.0 & 49.1 \\
1 & 0.0005 & 248 & 52.9 & 50.1 & 50.5 \\
1 & 0.001 & 156 & 48.8 & 49.0 & 49.1 \\
1 & 0.002 & 129 & 43.9 & 44.0 & 43.8 \\
2 & 0.0005 & 166 & 48.9 & 49.2 & 49.1 \\
2 & 0.001 & 127 & 43.1 & 43.3 & 43.2 \\
2 & 0.002 & 31 & 15.8 & 12.6 & 16.6 \\
\bottomrule
\end{tabular}
\end{table}

When target-derived controls are free, a fixed dense-control rung is hard to improve upon. Table~\ref{tab:select} in the appendix and Figure~\ref{fig:select}(a) replay the policies at $\nu = 0$, where a box costs its generation time alone and the five rungs cost within 18 s of one another. Fixed DK reaches 73.8\% in 185 s, the oracle 76.6\% with one generation, and the Pandora policy 74.5 to 75.1\% in 187 to 235 s, so the policy adds at most 1.3\% over the fixed rung for up to 27\% more time, because the oracle opens DK on 142 of the 148 clips and the remaining headroom is the choice of seed. The best of five seeds of the dense caption reaches 9.6\% for five generations per clip, and a random order with the same stopping rule falls to 63.1\%, so the prior supplies the whole gain over an uninformed search of the same boxes.

Cost-aware selection matters once obtaining or transmitting a dense control is expensive. Figure~\ref{fig:select}(b) prices the control at $\nu$ seconds per kB, and Table~\ref{tab:matched} compares three things at matched cost, the policy, the fixed-rung frontier that interpolates between neighboring fixed rungs at the control that the policy opened, and the per-clip routing rule at the same price. At $\nu = 0.5$, where the megabyte of a depth video costs as much as three generations, the policy reaches 65.7\% with 656 kB per clip, where the frontier gives 54.6\%, so the mixture of DK on the clips whose predicted gain pays for the megabyte and K2 or K1 on the others lies 11.1\% above the frontier at that budget. Elsewhere the policy lies above the frontier by 2.8 and 3.2\% at two points and within 1\% of it at the rest, so the gain at matched cost is concentrated where the price makes the dense rungs a per-clip decision. The routing rule reaches the same values within 1\% at every priced point except $\nu = 1$ and $\lambda = 0.0005$, where the second box of the policy adds 2.4\%, so the prior supplies the matched-cost gain and the reservation-value search adds a seed search that pays when information is cheap. A random order spends 1.2 to 3.0 times the control of the policy at the same price, and the greedy order by the prior mean ignores the price and opens DK on every clip.

Figure~\ref{fig:select}(c) replays the policies at $\nu = 0$ under critics of decreasing reliability, with noise of 10\% and 20\%, an appearance critic that sees DINO and CLIP alone, and a no-reference critic that never sees the reference~\cite{trustfl}. Uniform opening of all 13 boxes falls from 76.6\% to 73.9\%, 71.1\%, 65.5\% and 14.3\%, and the best of five seeds falls from 9.6\% to -0.9\%, below the fixed first-frame rung. The Pandora policy at $\lambda = 0.0005$ stays within 0.6\% of its value under the exact critic and pays only in budget, from 1.27 to 1.63 boxes per clip, because a critic that under-scores the first output fails to stop the search, while the prior decides which boxes open. The routing rule never consults the critic and keeps 73.7\% throughout. This collapse is what the audit of inference-time compute predicts for a weak verifier~\cite{computeaudit}.

\noindent\textbf{Human Audit.} We recruited 10 human raters to evaluate the composite critic on 128 pairs of outputs that share a reference clip, 16 from each of eight rung-pair types, stratified by the absolute composite-score difference. Raters viewed the reference beside two randomly positioned outputs and selected the output that better reproduced it overall, in appearance and in motion. The human majority agreed with the composite on 80.3\% of the pairs with an overall preference, above Struct at 78.7\% and DINO at 77.9\%. The checklist judge reached 57.5\%, and the aesthetic score, subject consistency and warping error reached 49.2\%, 39.3\% and 33.6\%, which matches the collapse of the no-reference critic in Figure~\ref{fig:select}(c). Raters differed on some close comparisons, with Fleiss' $\kappa=0.69$ on the overall question, and each rater agreed with the composite on 74.2 to 85.1\% of the pairs, as Section~\ref{sec:app-audit} in the appendix details.

\section{Limitations}
\label{sec:limitations}

The planner sees the target, so every closure is an empirical reference for a fixed student and sampler under privileged information, and a method that changes the architecture, the training or the sampler can exceed it. The clip features of the capability model are read from the reference, so the prior serves a user who holds the target, and a text-only deployment would need features predicted from the prompt. The dense-control rungs are measured on one student, since the second student accepts captions and a first frame only, so the transfer of depth and pose to other architectures is untested. The fidelity metrics and the checklist judge stand in for the match that a user perceives, and the human audit of Section~\ref{sec:selection} finds the composite close to the human majority and the checklist judge, which is least reliable on the order of events, near chance. The replay estimates policies on a grid generated in advance, the grid covers clips of about five seconds, and the released grid places a longer or a larger student on the same ladder without regenerating the references, since every control and score is included.

\section{Conclusion}
\label{sec:conclusion}

We measured how far a small video generator gets toward a target as it learns more about it, and we read the result as an empirical headroom reference for the methods that must produce the same information. A prompt leaves most of the appearance gap and almost all of the motion gap open, a keyframe closes appearance, a depth video closes motion, and the pair closes most gaps on the principal student across four teachers and 17 content dimensions. The capability model forecasts which clips a rung will close, and the replay shows that a fixed dense-control rung is hard to beat when controls are free, that the prior-driven policy lies above the fixed-rung frontier at matched cost once controls are priced, and that it survives a weak critic.

{
\small
\bibliographystyle{ieeenat_fullname}
\bibliography{refs}
}

\clearpage
\appendix
\twocolumn[{\begin{center}{\Large\bfseries Beyond Prompt Imitation: Capability-Aware\\Video Generation with Smaller Models}\par\vspace{0.8em}{\large Appendix}\par\end{center}\vspace{1.2em}}]
\flushbottom
\setlength{\dblfloatsep}{15.4pt plus 2pt minus 2pt}
\setlength{\dbltextfloatsep}{8pt plus 4pt minus 2pt}
\setlength{\textfloatsep}{8pt plus 4pt minus 2pt}
\setlength{\intextsep}{10pt plus 2pt minus 2pt}
\setlength{\floatsep}{8pt plus 4pt minus 2pt}

\section{Construction of the Reference Clips}
\label{sec:app-clips}

Every source video is cut into shots with a hard-cut detector that adds a color-histogram distance in HSV space to a mean pixel difference of downscaled consecutive frames. A cut is placed where the difference exceeds 0.25 and three times the median of the surrounding second, with at least half a second between cuts, and two source frames on each side of a cut are dropped to avoid blended transition frames. A shot of at least 5 s yields clips of 81 frames at 16 fps, which is the native length of the student, with as many evenly spread windows as fit when neighbors overlap by at most a quarter second, and a shot between 3 s and 5 s yields one clip of 49 frames, which the student also accepts. Frames are resampled to 16 fps by nearest-frame selection, which turns a 24 fps source into a 1-2-1-2 cadence of source frames. Optical flow is computed at a stride of two output frames and therefore always spans three source frames, so the motion metrics do not see the cadence. Each clip is resized and center-cropped to the closest of five aspect buckets of about 400k pixels (832$\times$480, 704$\times$528, 624$\times$624, 528$\times$704 and 480$\times$832), and the same resolution is used for generation, so no output is rescaled before scoring.

The user videos are 61 pieces from Seedance, Jimeng and Kling at 1088$\times$816 or 1280$\times$720. The 10 s single takes give one or two clips each, and the 20 s to 30 s multi-shot pieces give one clip per qualifying shot, for 148 clips in total. The teacher videos are the sampled-video release of VBench-2.0, from which we take one clip per video at the start of its first qualifying shot, so that each teacher contributes one clip per prompt slot. Of the 204 prompts, 195 have a clip from every one of the four teachers, and 10 videos fall out because their shots are shorter than 3 s. We replaced 16 prompt slots whose Veo 3 samples were truncated at the source and 8 slots whose prompt repeated an earlier slot, since a prompt can belong to several dimensions and the release repeats its videos, and 12 Veo 3 clips use the second or third sample of their prompt. All 204 prompts match the official VBench-2.0 prompt list. A Sora video that falls short of 81 frames by at most 0.7\% is resampled over its full duration so that its clip reaches 81 frames, which applies to 201 clips. The teacher set has 774 clips of 81 frames and 32 clips of 49 frames.

\section{Implementation Details}
\label{sec:app-impl}

\noindent\textbf{Refinement Baseline.} The ladder bounds method families, and one family is run in full so that a practical method can be placed against its bound. Table~\ref{tab:refine} reports the result, where R1 and R2 are the outputs of the two rounds, Best round is the best of the dense caption and the two rounds under the composite critic, and Best seed is the best of three seeds of the dense caption under the same critic. We reproduce the iterative prompt refinement loop of PhyT2V~\cite{phyt2v}, RAPO++~\cite{rapopp} and VISTA~\cite{vista} with the planner as the critic. A round shows the planner 10 timestamped frames of the reference and 10 of the student's output from the previous round, and asks it to rewrite the prompt in 80 to 120 words so that the next output matches the reference more closely, correcting every visible difference in the subjects, the setting, the lighting, the camera and the order of the events. Round 1 starts from the dense caption output, the student regenerates every rewritten prompt at the same seed, and the loop runs for two rounds on the 148 user clips. The loop spends three generations per clip, so we also report the best of its three candidates under the composite critic of Section~\ref{sec:method} in the main text, and the best of three seeds of the dense caption under the same critic, which spends the same compute without a critic in the loop.

\begin{table}[t]
\centering
\caption{Selection policies at $\nu = 0$ under the exact critic, with the average control opened per clip (kB).}
\label{tab:select}
\small
\setlength{\tabcolsep}{3pt}
\begin{tabular}{lrrrr}
\toprule
Policy & Value (\%) & Time (s) & Control (kB) & Boxes \\
\midrule
Fixed T2 & 0.0 & 174 & 0 & 1 \\
Fixed K1 & 30.0 & 187 & 74 & 1 \\
Fixed K2 & 49.0 & 180 & 149 & 1 \\
Fixed D & 58.3 & 192 & 984 & 1 \\
Fixed DK & 73.8 & 185 & 1058 & 1 \\
Best of 5 at T2 & 9.6 & 870 & 2 & 5 \\
Uniform rungs & 74.1 & 918 & 2267 & 5 \\
Uniform boxes & 76.6 & 2344 & 4684 & 13 \\
Routing $\lambda{=}0.0005$ & 73.7 & 185 & 1058 & 1 \\
Pandora $\lambda{=}0.0005$ & 75.1 & 235 & 1309 & 1.27 \\
Pandora $\lambda{=}0.001$ & 74.8 & 215 & 1232 & 1.16 \\
Pandora $\lambda{=}0.002$ & 74.5 & 187 & 1064 & 1.01 \\
Random $\lambda{=}0.002$ & 63.1 & 322 & 692 & 1.78 \\
Oracle & 76.6 & 184 & 1028 & 1 \\
\bottomrule
\end{tabular}
\end{table}

\noindent\textbf{Budgeted Replay.} The replay runs on the 148 user clips with the dense caption, K1, K2, D and DK as rungs, at five seeds of the dense caption, three seeds of K2 and DK and one seed of K1 and D, for 13 boxes per clip, and every box has been generated and scored in the grid. The time $t_r$ of a rung is its mean generation time over the grid, from 161 s for the dense caption to 187 s for depth, with 4 s of depth extraction added to D and DK, and $s_r$ is the measured size of its control, which averages 74 kB for K1, 149 kB for K2, 984 kB for D and 1058 kB for DK. The prior is a ridge regression of the composite closure on the clip features of Section~\ref{sec:capability} in the main text, fitted per rung on the five folds of that section, with a residual standard deviation of 9\% to 15\% per rung. We vary $\lambda$ from 0.0005 to 0.008 in units of the closure fraction per second and $\nu$ over 0, 0.1, 0.5, 1 and 2 s per kB. The critic that reveals a value is the exact composite closure, the composite with Gaussian noise of standard deviation 10\% or 20\%, the closure on appearance and CLIP alone, or a no-reference score that averages the aesthetic score on a unit scale with the subject consistency and never sees the reference. Every policy reports the true composite closure of the output that its critic prefers, averaged over clips, together with the time and the control that it is charged for every box that it opens, and Table~\ref{tab:select} lists every policy at $\nu = 0$, where the time of a rung is its mean generation time over the whole grid.

\begin{table}[t]
\centering
\caption{Prompt refinement against the ladder, gap closure (\%).}
\label{tab:refine}
\small
\setlength{\tabcolsep}{3pt}
\begin{tabular}{lrrrrrr}
\toprule
 & R1 & R2 & Best round & Best seed & T3 & K1 \\
\midrule
DINO & 5 & 3 & 9 & 3 & -11 & 53 \\
CLIP & 0 & 1 & 2 & 1 & -1 & 50 \\
Struct & 1 & 3 & 3 & 2 & -1 & 30 \\
Depth & 2 & 3 & 10 & 9 & -13 & 36 \\
Color & -2 & -1 & -1 & 1 & -4 & 55 \\
Flow & -2 & -4 & 5 & 6 & -5 & 8 \\
Camera & -5 & -9 & 5 & 7 & -12 & 5 \\
Object & -2 & -3 & 2 & 3 & -3 & 5 \\
Check & 13 & 17 & 37 & 35 & -3 & 61 \\
\midrule
Composite & 2 & 2 & 8 & 7 & -4 & 30 \\
\bottomrule
\end{tabular}
\end{table}

\section{Judge Validity}
\label{sec:app-judge}

\begin{table}[t]
\centering
\caption{Judge answers on the 954 reference clips by category.}
\label{tab:judge}
\small
\setlength{\tabcolsep}{4pt}
\begin{tabular}{lrrrr}
\toprule
 & \multicolumn{2}{c}{Yes-questions} & \multicolumn{2}{c}{No-questions} \\
\cmidrule(lr){2-3}\cmidrule(lr){4-5}
Category & Count & Missed (\%) & Count & Accepted (\%) \\
\midrule
Subject & 1676 & 2.2 & 567 & 8.6 \\
Action & 1378 & 9.7 & 1139 & 10.6 \\
Scene & 1097 & 1.9 & 617 & 3.6 \\
Style & 899 & 2.3 & 451 & 6.0 \\
Camera & 887 & 19.8 & 1048 & 2.9 \\
Count & 581 & 5.7 & 573 & 12.9 \\
Order & 330 & 27.0 & 148 & 12.8 \\
Other & 33 & 12.1 & 41 & 12.2 \\
\midrule
All & 6880 & 7.5 & 4584 & 7.6 \\
\bottomrule
\end{tabular}
\end{table}

The checklist judge is validated on the references themselves, where every yes-question should be confirmed and every no-question, which the planner writes as a plausible distractor, should be rejected. Table~\ref{tab:judge} reports the outcome on the 954 reference clips by question category. The judge confirms 92.5\% of the yes-questions and rejects 92.4\% of the distractors, so the two kinds of error are balanced and the checklist score, which averages the two rates, carries no yes bias. The errors concentrate in two categories. Camera questions are missed on 19.8\% of the references, since a slow dolly or a subtle zoom is hard to read from sampled frames, and temporal order questions are missed on 27.0\%, which is the category where a frame-sampling judge is weakest. Count questions are the main source of false confirmations, at 12.9\% of the distractors, because a distractor that changes the number of people or objects by one is confirmed when the judge miscounts. The per-question validity filter of Section~\ref{sec:setup} in the main text removes every question that the judge answers wrongly on its own reference, which leaves 951 of the 954 clips with at least two valid yes-questions and one valid no-question, and the checklist scores in the paper are computed on those questions only, so a question that the judge cannot answer on the reference is never charged against the student.

\section{Human Audit}
\label{sec:app-audit}

\begin{table}[t]
\centering
\caption{Agreement (\%) of the human majority with each score.}
\label{tab:audit}
\small
\setlength{\tabcolsep}{4pt}
\begin{tabular}{lrrr}
\toprule
Score & Overall & Appearance & Motion \\
\midrule
Composite & \textbf{80.3} & 78.8 & \textbf{82.1} \\
\midrule
DINO & 77.9 & \textbf{79.8} & 77.4 \\
CLIP & 69.7 & 70.2 & 62.3 \\
Struct & 78.7 & \textbf{79.8} & 74.5 \\
Depth & 70.5 & 68.3 & 67.9 \\
Color & 67.2 & 70.2 & 61.3 \\
Flow & 67.2 & 71.2 & 75.5 \\
Camera & 63.1 & 66.3 & 75.5 \\
Object & 66.4 & 67.3 & 73.6 \\
Check & 57.5 & 48.3 & 55.2 \\
\midrule
Aes & 49.2 & 53.8 & 45.3 \\
Subject consistency & 39.3 & 41.3 & 34.0 \\
Warping error & 33.6 & 30.8 & 34.0 \\
\midrule
Pairs with a preference & 122 & 104 & 106 \\
\bottomrule
\end{tabular}
\end{table}

\noindent\textbf{Protocol.} We sampled 128 pairs of outputs that share a reference clip from eight pair types, namely the dense caption against K1, D, the first refinement round and T3, K1 against DK and K2, D against DK, and two seeds of the dense caption. Each pair type contributed 16 pairs, stratified by the absolute composite-score difference with thresholds of 0.05, 0.15 and 0.3. Each trial looped the reference and the two outputs at a height of 288 pixels in random positions. All 10 raters independently answered left, same or right on all 128 pairs, without access to the rungs, the seeds or the scores, for three questions on overall fidelity, appearance fidelity (subjects, colors and scene) and motion fidelity (camera and subject motion). Responses were encoded as +1, -1 or 0, the majority preference was the sign of the summed votes, and a score agreed with the majority when the sign of its difference between the two outputs matched it, with the sign of the warping error reversed and ties excluded. Confidence intervals came from a bootstrap over the pairs. Inter-rater agreement was 84.6\% on the overall question, with Fleiss' $\kappa=0.69$, 82.1\% on appearance and 85.3\% on motion, and each rater agreed with the composite on 74.2 to 85.1\% of the pairs where that rater expressed a preference.

\noindent\textbf{Metric Agreement.} In Table~\ref{tab:audit}, the composite agreed with the overall majority on 80.3\% of the 122 pairs with a preference, with a bootstrap confidence interval of [73\%, 87\%]. It achieved the highest agreement on the overall and motion questions and stayed within 1.0\% of DINO and Struct on the appearance question, so it aligns with people on all three aspects of fidelity without relying on a single descriptor. The motion metrics aligned more closely with the motion question than with the overall question, since Flow, Camera and Object gained 7.2 to 12.4\% from the overall to the motion question, while CLIP and Color agreed less often on motion than on appearance, which supports separating appearance from temporal fidelity in the human evaluation. The checklist judge differed between the two outputs on only 73 of the 122 pairs, because many checklist questions receive the same answer for two outputs of one clip, and its agreement of 57.5\%, with an interval of [47\%, 68\%], covers chance. The no-reference scores aligned even less and fell at or below chance. The output that people preferred had the higher warping error on 66.4\% of the compared pairs, which is the tension between temporal smoothness and faithful reproduction that Section~\ref{sec:res-ladder} in the main text describes.

\noindent\textbf{Stratified Analysis.} Over the four strata of the absolute composite-score difference, the agreement was 87.5\%, 71.0\%, 80.0\% and 83.8\% from the closest to the clearest stratum, with 24 to 37 evaluated pairs per stratum, and the majority expressed no preference on 14\% of the closest pairs and on at most 3\% of the pairs in the other strata. The agreement does not rise monotonically with the score separation, and it stays high on both close and widely separated comparisons. The agreement also varied by pair type, from 100\% for D against DK to 68.8\% for K1 against K2 and for the two seeds of the dense caption, so the difficulty of a ranking depends on the conditions being compared as well as on the composite-score gap.

\noindent\textbf{Beyond Ladder Order.} On the 106 comparisons between different rungs, choosing the higher rung agreed with the human majority on 75.5\% of the pairs and choosing by the composite on 82.1\%. On the 19 comparisons where the composite preferred the lower rung, the majority agreed with the composite in 13 cases, so the critic identifies fidelity differences that the amount of control does not explain. DINO alone nearly matched the composite, and the replay of Figure~\ref{fig:select}(c) in the main text scores every policy with the exact composite, so the drop to 65.5\% under the appearance critic is a composite difference, and the audit cannot measure how much of it a viewer perceives.

\section{Sampling Steps}
\label{sec:app-steps}

\begin{table}[t]
\centering
\caption{Paired score difference of 50 minus 30 sampling steps per rung, and gap closure (\%) within each step count, on 20 user clips.}
\label{tab:steps}
\small
\setlength{\tabcolsep}{3pt}
\begin{tabular}{lrrrrrrr}
\toprule
 & \multicolumn{3}{c}{Difference} & \multicolumn{2}{c}{K1 closure} & \multicolumn{2}{c}{DK closure} \\
\cmidrule(lr){2-4}\cmidrule(lr){5-6}\cmidrule(lr){7-8}
 & T2 & K1 & DK & 30 & 50 & 30 & 50 \\
\midrule
DINO & +0.045 & +0.008 & -0.007 & 57 & 54 & 77 & 73 \\
CLIP & +0.001 & -0.004 & +0.002 & 46 & 45 & 60 & 60 \\
Struct & -0.008 & -0.001 & -0.003 & 25 & 25 & 65 & 65 \\
Depth & -0.044 & +0.003 & -0.002 & 25 & 34 & 93 & 93 \\
Color & +0.060 & +0.019 & +0.006 & 54 & 53 & 61 & 57 \\
Flow & -0.030 & -0.040 & -0.005 & 16 & 16 & 81 & 82 \\
Camera & -0.037 & -0.057 & -0.003 & 17 & 16 & 83 & 84 \\
Object & -0.044 & -0.032 & -0.008 & 7 & 12 & 71 & 72 \\
Check & +0.018 & -0.002 & +0.001 & 60 & 62 & 67 & 66 \\
Aes & +0.011 & +0.091 & -0.037 & -- & -- & -- & -- \\
\bottomrule
\end{tabular}
\end{table}

Every output in the paper is sampled with 30 UniPC steps, and the choice is validated on 20 random user clips at the dense caption, first-frame and depth-with-first-frame rungs, which are generated again at seed 0 with 50 steps and scored in the same way, for 60 paired outputs. Table~\ref{tab:steps} reports the paired difference of every criterion between the two step counts and the closure of the two visual rungs over the dense caption within each step count. The longer schedule changes every criterion of every rung by less than 0.07, and the bootstrap interval of the difference covers zero on 24 of the 27 pairs of fidelity criterion and rung. The three exceptions are small. The dense caption gains 0.045 in appearance and 0.060 in color at 50 steps, and the depth rung loses 0.007 in appearance. The motion criteria of the dense caption and of the first frame fall by 0.03 to 0.06 with intervals that cover zero, the first frame gains 0.09 in aesthetic score on its 10-point scale, and the depth rung moves by less than 0.01 on every fidelity criterion, since the dense control leaves the sampler little freedom.

The closures that the paper reads from the ladder are stable as well. The first frame closes 57\% of the appearance gap at 30 steps and 54\% at 50, depth with the first frame closes 77\% and 73\%, the motion closures of the two rungs agree within 1\%, and the depth closure of DK is 93\% at both step counts. The largest change is the object motion closure of K1, from 7\% to 12\%, whose interval at 30 steps spans \mbox{[-8, 17]\%} on 20 clips, and the depth closure of K1, from 25\% to 34\%, with an interval of \mbox{[7, 42]\%} at 30 steps. A longer schedule therefore moves the student by a few hundredths of a score and leaves the closure of each rung and the ordering of the rungs unchanged, so the 30 step grid supports the conclusions of the paper at 60\% of the sampling cost of the 50 step default.

\section{Second Student}
\label{sec:app-profile}

\begin{table}[t]
\centering
\caption{Dense caption scores and gap closure (\%) of the caption and keyframe rungs on the user clips for both students.}
\label{tab:wan22}
\small
\setlength{\tabcolsep}{4pt}
\begin{tabular}{lrrrrrrr}
\toprule
 & \multicolumn{4}{c}{Wan2.2-TI2V-5B} & \multicolumn{3}{c}{VACE-1.3B} \\
\cmidrule(lr){2-5}\cmidrule(lr){6-8}
 & T2 score & T1 & T3 & K1 & T1 & T3 & K1 \\
\midrule
DINO & 0.56 & -43 & -13 & 51 & -39 & -11 & 53 \\
CLIP & 0.69 & -22 & -4 & 51 & -20 & -1 & 50 \\
Struct & 0.31 & -7 & -2 & 30 & -5 & -1 & 30 \\
Depth & 0.47 & -11 & -9 & 37 & -14 & -13 & 36 \\
Color & 0.38 & -14 & -2 & 55 & -9 & -4 & 55 \\
Flow & 0.19 & -7 & -2 & 13 & -9 & -5 & 8 \\
Camera & 0.21 & -11 & -4 & 12 & -17 & -12 & 5 \\
Object & 0.16 & -2 & -1 & 8 & -5 & -3 & 5 \\
Check & 0.88 & -87 & -14 & 50 & -99 & -3 & 61 \\
\bottomrule
\end{tabular}
\end{table}

\begin{figure}[t]
\centering
\includegraphics[width=\columnwidth]{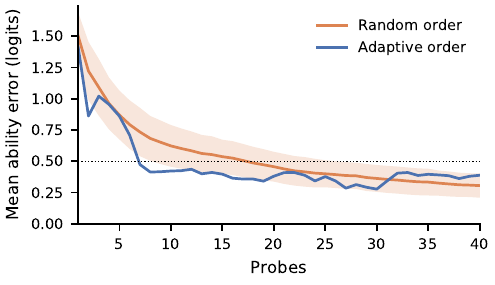}
\caption{K1 ability error of Wan2.2 against probed clips.}
\label{fig:profile}
\end{figure}

The capability model also profiles the second student from few generations. With the clip difficulties of the first student held fixed, the K1 abilities of Wan2.2 fitted on all 148 clips lie within 0.4 logits of those of VACE on every criterion, which restates the agreement of Table~\ref{tab:wan22} in the units of the model. Adaptive probing then recovers that profile from a handful of clips. Each probe generates one video at K1 and yields one outcome per criterion, the next clip is the one with the largest expected information gain summed over the eight abilities under the current posteriors, and the estimate after every probe is compared with the fit on the full grid. Figure~\ref{fig:profile} shows that the adaptive order brings the mean ability error below 0.5 logits after 7 probes, where a random order of clips needs 18, and that the errors after 10 probes are 0.42 and 0.62 logits. The two orders meet near 40 probes, where the random order has covered the range of difficulties as well, so the gain of adaptive probing is largest exactly where a user would stop, after a few generations.

\section{Capability Model Weights}
\label{sec:app-weights}

\begin{table}[t]
\centering
\caption{Difficulty weights of the standardized clip features in the capability model at half closure.}
\label{tab:weights}
\scriptsize
\setlength{\tabcolsep}{1.5pt}
\begin{tabular}{lrrrrrrrr}
\toprule
Feature & DINO & CLIP & Struct & Depth & Flow & Camera & Object & Check \\
\midrule
Caption words & -0.12 & -0.31 & -0.20 & -0.29 & 0.02 & 0.21 & 0.32 & -0.11 \\
Camera motion & -0.29 & -0.13 & -1.25 & -0.53 & 0.86 & 0.82 & 0.23 & 0.18 \\
Object motion & -0.30 & -0.38 & 0.10 & 0.40 & -1.53 & -1.49 & -0.42 & -0.07 \\
Number of subjects & -0.19 & 0.11 & -0.10 & 0.31 & -0.03 & -0.08 & 0.37 & 0.24 \\
Person present & 0.02 & 0.01 & 0.52 & -0.08 & -0.15 & -0.18 & 0.01 & -0.12 \\
Reference aesthetic & 0.04 & 0.56 & 0.30 & 0.07 & -0.01 & -0.14 & 0.62 & 0.01 \\
Reference warping error & -0.01 & -0.54 & -0.51 & -0.51 & 0.13 & 0.19 & 0.24 & -0.24 \\
Short clip & 0.07 & 0.07 & 0.03 & 0.34 & -0.17 & 0.03 & 0.09 & 0.16 \\
\bottomrule
\end{tabular}
\end{table}

Table~\ref{tab:weights} lists the weights $\mathbf{w}_k$ of Equation~\eqref{eq:irt} in the main text for every criterion, fitted on the 148 user clips at $\tau = 0.5$. A positive weight raises the difficulty of a clip, so the rung is less likely to close half of the gap on clips where the feature is large. The features are standardized over the clips, so the weights are comparable within a column. Camera motion and object motion are the mean magnitudes of the camera and residual parts of the reference flow in units of frame width, the number of subjects comes from the planner's program, person present is the fraction of frames with a detected person, the reference aesthetic score and warping error are the no-reference scores of the reference itself, caption words is the length of the dense caption, and short clip marks the clips of 49 frames. A teacher indicator is part of the feature set for the teacher clips and is constant on the user clips, so its weight is zero here and it is left out of the table. The two motion magnitudes carry the largest weights, as discussed in Section~\ref{sec:capability} in the main text, and the reference warping error is the next largest and lowers the difficulty on CLIP, structure and depth, which is consistent with the gap being relative, since a reference with more frame-to-frame change leaves a larger gap at the caption rung for the visual rungs to close.

\section{Capability Model on All Clips}
\label{sec:app-irtall}

\begin{table}[t]
\centering
\caption{Difficulty weights of the standardized clip features in the capability model at half closure on all 954 clips.}
\label{tab:weightsall}
\scriptsize
\setlength{\tabcolsep}{1.5pt}
\begin{tabular}{lrrrrrrrr}
\toprule
Feature & DINO & CLIP & Struct & Depth & Flow & Camera & Object & Check \\
\midrule
Caption words & -0.14 & -0.08 & -0.05 & -0.09 & 0.11 & 0.08 & 0.09 & -0.06 \\
Camera motion & -0.30 & -0.31 & -1.22 & -0.52 & 0.55 & 0.59 & -0.03 & -0.04 \\
Object motion & 0.06 & -0.01 & 0.15 & 0.28 & -0.51 & -0.64 & 0.11 & 0.07 \\
Number of subjects & 0.16 & 0.07 & 0.08 & 0.02 & 0.11 & 0.22 & 0.07 & 0.09 \\
Person present & 0.17 & 0.07 & 0.31 & 0.07 & 0.21 & 0.15 & 0.36 & 0.07 \\
Reference aesthetic & 0.06 & 0.19 & 0.02 & 0.11 & 0.11 & 0.09 & 0.06 & -0.05 \\
Reference warping error & -0.34 & -0.42 & -0.56 & -0.52 & -0.14 & -0.07 & -0.11 & -0.10 \\
Short clip & -0.03 & 0.02 & 0.11 & 0.12 & -0.07 & -0.02 & -0.05 & -0.02 \\
Teacher clip & -0.20 & 0.05 & 0.07 & 0.06 & -0.07 & -0.04 & 0.19 & 0.00 \\
\bottomrule
\end{tabular}
\end{table}

\begin{table}[t]
\centering
\caption{Held-out AUC of the success predictors and fitted rung abilities at half closure on all 954 clips.}
\label{tab:irtall}
\small
\setlength{\tabcolsep}{4pt}
\begin{tabular}{lrrrrrrr}
\toprule
 & \multicolumn{3}{c}{AUC} & \multicolumn{4}{c}{Rung ability $\theta_{rk}$} \\
\cmidrule(lr){2-4}\cmidrule(lr){5-8}
 & Rate & Rasch & GBM & K1 & K2 & D & DK \\
\midrule
DINO & 0.74 & 0.79 & 0.80 & 0.5 & 1.8 & -0.3 & 2.6 \\
CLIP & 0.83 & 0.88 & 0.87 & 0.5 & 1.9 & -2.4 & 2.0 \\
Struct & 0.77 & 0.87 & 0.91 & -1.2 & 0.5 & -1.0 & 3.1 \\
Depth & 0.82 & 0.87 & 0.88 & -0.1 & 1.1 & 3.9 & 4.6 \\
Flow & 0.89 & 0.91 & 0.94 & -3.4 & -1.2 & 1.7 & 2.1 \\
Camera & 0.86 & 0.90 & 0.92 & -2.9 & -0.8 & 1.6 & 1.9 \\
Object & 0.86 & 0.88 & 0.92 & -3.8 & -1.9 & 0.8 & 1.3 \\
Check & 0.60 & 0.60 & 0.58 & 0.8 & 1.6 & 0.7 & 1.6 \\
\bottomrule
\end{tabular}
\end{table}

Table~\ref{tab:irtall} reports the capability model of Equation~\eqref{eq:irt} in the main text fitted on all 954 clips at $\tau = 0.5$, and Table~\ref{tab:weightsall} lists the difficulty weights of the standardized features, now with the teacher indicator. The fit uses 3816 clip-rung observations per fidelity criterion and 2564 on the checklist, since a clip whose caption already satisfies every validated question leaves no checklist gap to close, and Section~\ref{sec:teacher} in the main text discusses what the two tables show about the teacher clips and the content dimensions.

\section{Teacher Clips and Content Dimensions}
\label{sec:app-teacher}

\begin{table*}[t]
\centering
\caption{Gap closure (\%) of three rungs over the dense caption on all clips and by source of the reference.}
\label{tab:teacher}
\small
\setlength{\tabcolsep}{4pt}
\begin{tabular}{lrrrrrrrrrrrrrrrrrr}
\toprule
 & \multicolumn{3}{c}{All} & \multicolumn{3}{c}{User} & \multicolumn{3}{c}{Kling} & \multicolumn{3}{c}{Sora} & \multicolumn{3}{c}{Veo 3} & \multicolumn{3}{c}{Vidu Q1} \\
\cmidrule(lr){2-4}\cmidrule(lr){5-7}\cmidrule(lr){8-10}\cmidrule(lr){11-13}\cmidrule(lr){14-16}\cmidrule(lr){17-19}
 & K1 & D & DK & K1 & D & DK & K1 & D & DK & K1 & D & DK & K1 & D & DK & K1 & D & DK \\
\midrule
DINO & 55 & 42 & 76 & 53 & 50 & 78 & 59 & 38 & 77 & 56 & 42 & 74 & 57 & 44 & 77 & 49 & 40 & 73 \\
CLIP & 54 & 22 & 67 & 50 & 21 & 63 & 56 & 22 & 69 & 56 & 22 & 66 & 58 & 24 & 69 & 50 & 22 & 64 \\
Struct & 39 & 44 & 72 & 30 & 47 & 68 & 43 & 42 & 74 & 39 & 44 & 70 & 48 & 45 & 75 & 34 & 44 & 69 \\
Depth & 42 & 90 & 94 & 36 & 91 & 95 & 46 & 87 & 95 & 41 & 91 & 95 & 51 & 91 & 90 & 35 & 91 & 94 \\
Color & 65 & 5 & 70 & 55 & 7 & 61 & 70 & 4 & 74 & 68 & 4 & 71 & 68 & 10 & 74 & 63 & 4 & 66 \\
Flow & 9 & 70 & 73 & 8 & 73 & 76 & 8 & 72 & 75 & 13 & 71 & 74 & 7 & 63 & 68 & 9 & 71 & 74 \\
Camera & 9 & 72 & 75 & 5 & 75 & 77 & 10 & 74 & 79 & 14 & 73 & 76 & 8 & 64 & 70 & 8 & 75 & 76 \\
Object & 8 & 58 & 64 & 5 & 55 & 62 & 6 & 56 & 62 & 12 & 58 & 65 & 8 & 59 & 65 & 9 & 61 & 64 \\
Check & 54 & 51 & 76 & 61 & 51 & 69 & 55 & 55 & 75 & 56 & 57 & 78 & 50 & 54 & 75 & 50 & 42 & 79 \\
\bottomrule
\end{tabular}
\end{table*}

\begin{table*}[t]
\centering
\caption{Gap closure (\%) of three rungs over the dense caption on the teacher clips by VBench-2.0 dimension.}
\label{tab:dimension}
\small
\setlength{\tabcolsep}{3pt}
\begin{tabular}{lrrrrrrrrrrrrrrrr}
\toprule
\multirow{2}{*}{Dimension} & \multirow{2}{*}{Clips} & \multicolumn{3}{c}{DINO} & \multicolumn{3}{c}{Depth} & \multicolumn{3}{c}{Flow} & \multicolumn{3}{c}{Camera} & \multicolumn{3}{c}{Check} \\
\cmidrule(lr){3-5}\cmidrule(lr){6-8}\cmidrule(lr){9-11}\cmidrule(lr){12-14}\cmidrule(lr){15-17}
 & & K1 & D & DK & K1 & D & DK & K1 & D & DK & K1 & D & DK & K1 & D & DK \\
\midrule
Camera Motion & 48 & 66 & 28 & 76 & 40 & 93 & 96 & 16 & 72 & 77 & 20 & 77 & 82 & 53 & 49 & 67 \\
Complex Landscape & 47 & 51 & 28 & 67 & 19 & 75 & 65 & 7 & 65 & 69 & 1 & 67 & 71 & 77 & 25 & 74 \\
Complex Plot & 44 & 49 & 41 & 71 & 31 & 88 & 92 & 9 & 68 & 72 & 10 & 72 & 76 & 19 & 46 & 67 \\
Composition & 47 & 52 & 43 & 75 & 32 & 91 & 95 & 4 & 74 & 76 & 6 & 76 & 77 & 52 & 57 & 82 \\
Dynamic Attribute & 47 & 64 & 29 & 76 & 41 & 79 & 91 & 9 & 53 & 65 & 11 & 53 & 63 & 57 & 52 & 72 \\
Dynamic Spatial Relationship & 47 & 35 & 34 & 66 & 50 & 94 & 97 & 6 & 76 & 78 & 8 & 80 & 82 & 43 & 43 & 50 \\
Human Anatomy & 48 & 52 & 54 & 81 & 45 & 96 & 98 & 17 & 83 & 84 & 18 & 86 & 86 & 43 & 53 & 80 \\
Human Clothes & 48 & 59 & 50 & 83 & 49 & 95 & 97 & 16 & 76 & 80 & 16 & 79 & 83 & 57 & 56 & 82 \\
Human Identity & 48 & 50 & 56 & 80 & 34 & 94 & 97 & 11 & 79 & 80 & 13 & 83 & 84 & 41 & 72 & 85 \\
Human Interaction & 48 & 53 & 60 & 82 & 42 & 96 & 98 & 11 & 82 & 83 & 13 & 88 & 88 & 69 & 78 & 90 \\
Instance Preservation & 48 & 48 & 28 & 68 & 35 & 89 & 93 & 6 & 63 & 67 & 4 & 60 & 67 & 50 & 59 & 82 \\
Material & 48 & 61 & 34 & 70 & 48 & 82 & 93 & 3 & 49 & 53 & 4 & 56 & 58 & 58 & 41 & 74 \\
Mechanics & 48 & 56 & 32 & 75 & 52 & 82 & 93 & 8 & 58 & 65 & 11 & 53 & 61 & 60 & 41 & 80 \\
Motion Order Understanding & 48 & 39 & 38 & 66 & 52 & 94 & 97 & 5 & 80 & 80 & 3 & 85 & 84 & 43 & 53 & 79 \\
Motion Rationality & 48 & 63 & 51 & 84 & 54 & 96 & 99 & 10 & 77 & 81 & 12 & 81 & 83 & 72 & 32 & 77 \\
Multi-View Consistency & 48 & 68 & 51 & 83 & 50 & 94 & 98 & 15 & 79 & 82 & 12 & 74 & 78 & 59 & 74 & 89 \\
Thermotics & 46 & 68 & 40 & 75 & 59 & 85 & 94 & 8 & 37 & 44 & 9 & 43 & 53 & 55 & 43 & 83 \\
\bottomrule
\end{tabular}
\end{table*}

Table~\ref{tab:teacher} runs the ladder on all 954 clips at the dense caption, first-frame, depth and depth-with-first-frame rungs, with the two-keyframe rung generated on the same clips, and breaks the closure down by the source of the reference, which is the 148 user clips or the 204, 201, 201 and 200 clips of Kling, Sora, Veo 3 and Vidu Q1. On all clips, the dense caption reaches a DINO similarity of 0.57, a structure similarity of 0.34 and a flow agreement of 0.11, the first frame closes 55\% of the appearance gap and 9\% of the flow gap, two keyframes close 70\% and 30\%, the depth video closes 70\% of the flow gap and 5\% of the color gap, and depth with the first frame closes 64 to 94\% of every fidelity gap, with bootstrap intervals of at most 3\% on either side on the fidelity axes, so the shape that Section~\ref{sec:res-ladder} in the main text read from the user clips is the shape of the whole grid. The four teachers agree with one another on nearly every cell. The first frame closes 49 to 59\% of the appearance gap and 7 to 13\% of the flow gap, the depth video closes 63 to 72\% of the flow gap and 4 to 10\% of the color gap, and the pair closes 73 to 77\% of the appearance gap and 90 to 95\% of the depth gap, so the order of the rungs on every axis is the same for every teacher. Two differences stand out. The teacher clips leave a larger color gap at the dense caption than the user clips, with a score of 0.22 to 0.31 against 0.38, and the first frame closes more of it, 63 to 70\% against 55\%, so the color of a teacher clip is less often implied by its caption and more often fixed by its first frame. Veo 3 is the teacher on which the depth video hands over the least motion, with 63\% of the flow gap and 64\% of the camera gap against 71 to 75\% on the other three, and its depth closure under DK is 90\% with an interval of [79, 96]\% where the other three are within 1\% of 95, which points to a few Veo 3 clips whose motion the student does not follow from the depth video. Vidu Q1 has the lowest first-frame closure on appearance and color, 49\% and 63\%, and the lowest depth closure of the checklist, 42\% with an interval of [28, 54]\% that overlaps the intervals of the other teachers.

Table~\ref{tab:dimension} breaks the teacher clips down by the VBench-2.0 dimension of their prompt, with each prompt assigned to one dimension, which gives 17 dimensions of 44 to 48 clips. The dimension moves the closure of a rung far more than the teacher does, and it moves it along the axis that the dimension names. The depth video hands over the least motion on Thermotics, where it closes 37\% of the flow gap and 43\% of the camera gap, on Material with 49 and 56\%, on Dynamic Attribute with 53 and 53\% and on Mechanics with 58 and 53\%, which are the dimensions whose prompts describe melting, burning, deformation and a change of material or attribute, so the motion of the reference is a change of the surface itself and a depth map carries little of it, and even depth with the first frame closes only 44\% of the flow gap on Thermotics. The depth video hands over the most motion on the human dimensions, with 82 to 83\% of the flow gap and 86 to 88\% of the camera gap on Human Anatomy and Human Interaction, and on Motion Order Understanding with 80 and 85\%, where a body moves through a scene that the depth map describes well. The first frame closes the least of the appearance gap on Dynamic Spatial Relationship with 35\% and Motion Order Understanding with 39\%, whose clips rearrange their layout over time, and the most on Multi-View Consistency, Thermotics and Camera Motion with 66 to 68\%, where the same subject stays in view throughout. Complex Landscape is the one dimension where depth with the first frame closes less than 90\% of the depth gap, 65\% with an interval of [21, 90]\%, because its wide scenes already score 0.79 on depth at the dense caption against 0.29 to 0.62 elsewhere, so the closure is a ratio of two small quantities and its interval is wide. The checklist closures carry the widest intervals, since the caption already satisfies 88 to 95\% of the questions on every dimension, and the first frame closes 19\% of the remaining checklist gap on Complex Plot with an interval that spans zero, while depth with the first frame closes 67 to 90\% of it on every dimension except Dynamic Spatial Relationship, where it closes 50\% of the gap with an interval of [-7, 87]\% that covers most of the scale.

The capability model keeps its profile and its accuracy when it is refitted on all 954 clips. Table~\ref{tab:irtall} reports the fit on 3816 clip-rung observations per fidelity criterion, with the features of Section~\ref{sec:capability} in the main text plus an indicator of a teacher clip and the same five-fold cross-validation over clips. The Rasch model raises the held-out AUC over the per-rung rate by 0.02 to 0.10 on the seven fidelity criteria and reaches 0.87 to 0.91 on structure, depth, flow and camera, and the boosted classifier is within 0.04 of it everywhere, so the clip features explain the closure of a teacher clip as well as that of a user clip. The rung abilities keep the order that Section~\ref{sec:capability} in the main text reports on the user clips, with depth with the first frame above 2 on appearance, CLIP, structure, depth and flow, depth alone at 3.9 on depth and -2.4 on CLIP, and the first frame below -2.9 on flow, camera and object motion. The teacher indicator has its largest weights at -0.20 on appearance and 0.19 on object motion, so a teacher clip is slightly easier on appearance and slightly harder on object motion than a user clip with the same features, and the source of a clip adds little to what its features already say. Camera motion keeps the largest weight, -1.2 on structure, while the weights of object motion on flow and camera, -0.5 and -0.6, are a third to a half of their size on the user clips, and Table~\ref{tab:weightsall} lists every weight. The checklist stays unpredictable from the features, with an AUC of 0.60 for the per-rung rate and for the model alike, as on the user clips, because the dense caption already satisfies most checklist questions for every source.

\section{Sufficiency Profile}
\label{sec:app-suff}

\begin{figure*}[t]
\centering
\includegraphics[width=\textwidth]{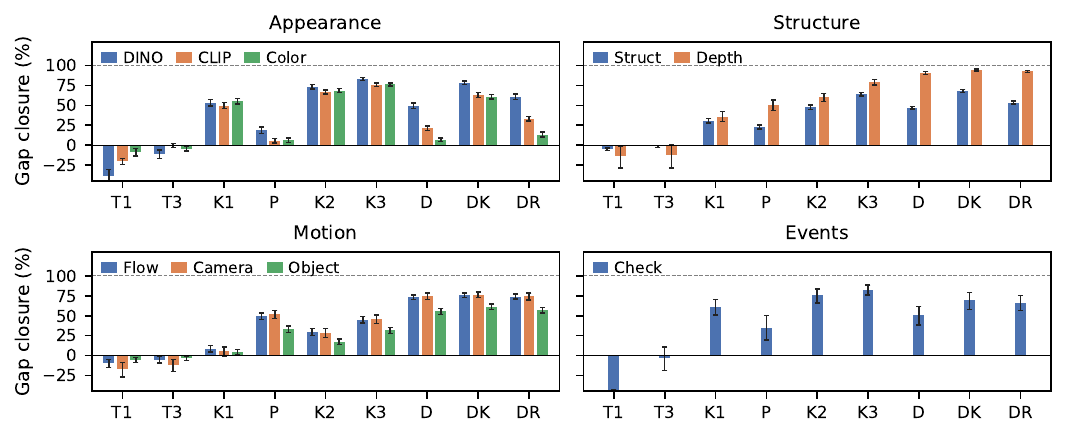}
\caption{Gap closure (\%) of the other rungs over the dense caption on the user clips, with bootstrap intervals over clips.}
\label{fig:ladder}
\end{figure*}

\begin{table}[t]
\centering
\caption{Lowest sufficient rung at half closure (\% of user clips).}
\label{tab:suff}
\small
\setlength{\tabcolsep}{3pt}
\begin{tabular}{lrrrrrrrrrrr}
\toprule
 & T1 & T2 & T3 & K1 & P & K2 & K3 & D & DK & DR & None \\
\midrule
DINO & 1 & 0 & 3 & 59 & 1 & 25 & 9 & 1 & 1 & 0 & 1 \\
CLIP & 0 & 0 & 0 & 51 & 1 & 32 & 14 & 0 & 0 & 0 & 2 \\
Struct & 0 & 0 & 0 & 18 & 0 & 22 & 43 & 3 & 5 & 0 & 8 \\
Depth & 10 & 0 & 3 & 25 & 22 & 21 & 14 & 5 & 0 & 0 & 1 \\
Color & 0 & 0 & 0 & 65 & 0 & 22 & 9 & 0 & 1 & 0 & 3 \\
Flow & 0 & 0 & 0 & 4 & 34 & 11 & 13 & 32 & 2 & 1 & 4 \\
Camera & 1 & 0 & 1 & 7 & 36 & 12 & 10 & 24 & 3 & 0 & 6 \\
Object & 0 & 0 & 0 & 1 & 17 & 4 & 9 & 38 & 8 & 1 & 22 \\
Check & 10 & 21 & 18 & 34 & 4 & 6 & 3 & 0 & 0 & 0 & 3 \\
\bottomrule
\end{tabular}
\end{table}

Figure~\ref{fig:ladder} draws the closures of Table~\ref{tab:ladder} in the main text with their bootstrap intervals over clips, and these averages hide that clips differ in what they need. Table~\ref{tab:suff} reports, for every criterion, the share of clips whose lowest sufficient rung is each rung of the ladder, where a rung is sufficient when it closes at least half of the gap, and the rungs are walked in the order of control size of Table~\ref{tab:rungs} in the main text, so a clip is credited to the cheapest rung that works and a clip without a person skips the pose rung. A first frame is enough for appearance on 51 to 65\% of the clips across DINO, CLIP and color, and the keyframe rungs together reach 93 to 97\%, so the appearance of almost every clip is a matter of keyframes. Motion is the mirror image. Keyframes alone reach half closure on 28 to 29\% of the clips for flow and camera motion, pose is the lowest sufficient rung on 34 to 36\%, since the skeleton is cheaper than the depth video, and depth on another 24 to 32\%. No rung satisfies object motion on 22\% of the clips, which is the capacity limit of the student. On the checklist, 21\% of the clips are already at the ceiling with the dense caption, the short caption or the program reaches half closure on another 28\%, and the first frame is the lowest sufficient rung for 34\%.

The profile is a statement about the student, so it should be stable along the chains of the ladder where each rung contains the information of the one below. Along the depth chain T2, D, DK, closure never falls by more than 0.05 when climbing on 88 to 100\% of the clips for every fidelity criterion but color, where the share is 81\%, and on 78\% for the checklist. Along the keyframe chain T2, K1, K2, K3, the share is 92 to 97\% for appearance, CLIP, structure and color, and it falls to 55 to 70\% for the three motion criteria, so adding a keyframe moves the motion of the output away from the reference on a third to a half of the clips even though it improves the mean. The student interpolates between fixed frames along its own path, an instability that the mean flow closure in Table~\ref{tab:ladder} in the main text conceals, while the depth video, which fixes the path, climbs monotonically on 95\% of the clips for flow and object motion.

\section{Planner and Judge Protocols}
\label{sec:app-prompts}

\begin{figure*}[t]
\centering
\includegraphics[width=\textwidth]{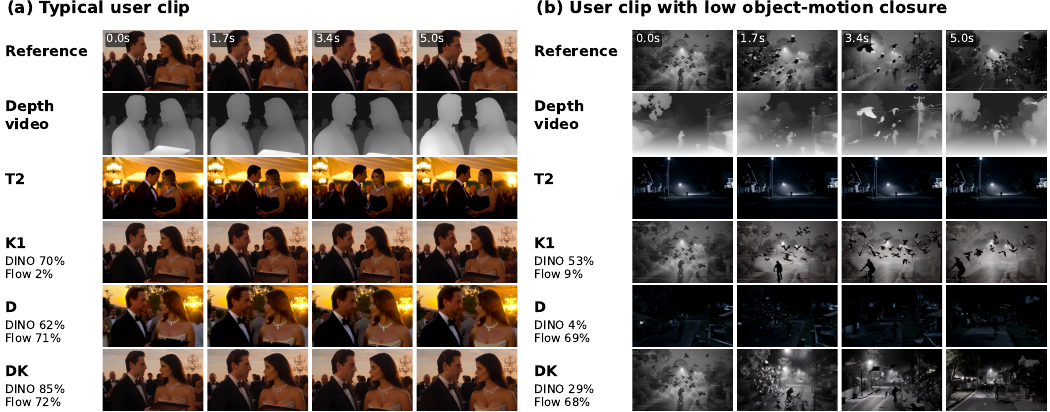}
\caption{Depth video and outputs of the student on two user clips, with the DINO and flow closures of each rung.}
\label{fig:examples-user}
\end{figure*}

\begin{figure*}[t]
\centering
\includegraphics[width=\textwidth]{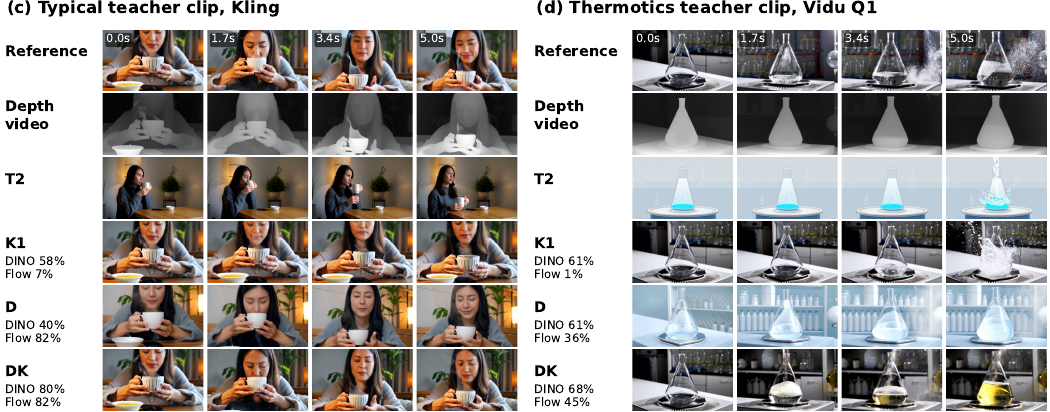}
\caption{Depth video and outputs of the student on two teacher clips, with the DINO and flow closures of each rung.}
\label{fig:examples-teacher}
\end{figure*}

The planner receives 12 frames sampled evenly from the clip, each preceded by its timestamp, together with the duration of the shot, and it is told to describe only what is visible, with no guessed names or story, and to ignore burned-in text overlays. It returns one JSON object whose fields are the three text rungs and the clip features. The short caption is one sentence of at most 25 words, and the dense caption is one paragraph of 80 to 120 words written as a text-to-video prompt that covers the subjects and their appearance, the actions in temporal order, the scene, the camera framing and movement, the lighting and the style. The structured program of rung T3 renders the remaining fields, which are the scene as a phrase, the subjects as a list with an identifier, a category and an appearance, the camera as a shot type, a motion type and a motion strength, the timeline as two to five events with start and end times and one action clause each, and the style as a look and a lighting. A second pass of the planner writes the checklist of 10 to 12 questions that a careful viewer could answer with yes or no from the video alone. Seven or eight questions have the answer yes, three or four have the answer no and describe a plausible alternative that the video does not show, such as a reversed order or a different count, and every question carries one of the categories of Table~\ref{tab:judge}. The judge receives the frames of an output with their timestamps and the numbered questions of the clip, with the instruction to use only what the frames show and to ignore burned-in text overlays, and it replies with one line per question that carries the number and yes or no. The same call on the reference clip gives the validity filter of Section~\ref{sec:setup} in the main text.
\section{Qualitative Examples}
\label{sec:app-examples}

Figures~\ref{fig:examples-user} and \ref{fig:examples-teacher} show the reference, the depth video that D and DK receive, and the outputs of T2, K1, D and DK at four timesteps on four clips. The clips follow a fixed rule. Each is the clip whose closures of K1, D and DK on appearance, flow, color and object motion lie closest to the median of its pool, where the pools are all user clips, the 33 user clips on which no rung closes half of the object-motion gap, all teacher clips and the Thermotics teacher clips, without the clip of Figure~\ref{fig:teaser} in the main text.

On the typical clips (a) and (c), the rungs behave as the averages of Table~\ref{tab:ladder} in the main text predict. The dense caption draws the named content in its own scene, lighting and framing. The first frame fixes the look of the scene and the subjects but leaves them almost still, with flow closures of 2\% and 7\%. The depth video reproduces the motion, with flow closures of 71\% and 82\%, but repaints the scene, so the couple of panel (a) gains warm light and the person of panel (c) a different face and room. DK keeps both, with 72 to 85\% closure on appearance and flow. Panel (b) shows a night street through which many small objects fly. The depth video carries the objects as bright blobs, but D renders the street almost black, with 4\% appearance closure, while K1 keeps the look of the street but moves it little, and no rung closes half of the object-motion gap. Panel (d) shows a Thermotics prompt in which vapor rises from a heated flask. The depth video keeps the flask and loses the vapor, a change of the material that depth does not encode, so D and DK close 36\% and 45\% of the flow gap, the lowest values among the four clips.

\par\penalty-10000

\end{document}